\documentclass{ifacconf}

\usepackage{graphicx}      
\usepackage{natbib}        
\usepackage{epsfig} 
\usepackage{subfigure}
\usepackage{amsmath} 
\usepackage{amssymb}  
\usepackage{makecell}
\usepackage{multirow}
\begin{document}
\begin{frontmatter}

\title{A Multi-view Landmark Representation Approach with Application to GNSS-Visual-Inertial Odometry} 

\thanks[footnoteinfo]{This work was supported by the National Natural Science Foundation (62303310).}

\author[First]{Tong Hua} 
\author[First]{Jiale Han} 
\author[Second]{Wei Ouyang}

\address[First]{ School of Electronic Information and Electrical Engineering, 
   Shanghai Jiao Tong University, Shanghai, 200240, China (e-mail: \{ht1234,hanjl2022\}@sjtu.edu.cn).}
\address[Second]{College of Surveying and Geo-Informatics, Tongji University, Shanghai, 200092, China (Corresponding author, e-mail: ywoulife@tongji.edu.cn)}

\begin{abstract}
Invariant Extended Kalman Filter (IEKF) has been a significant technique in vision-aided sensor fusion. However, it usually suffers from high computational burden when jointly optimizing camera poses and the landmarks. To improve its efficiency and applicability for multi-sensor fusion, we present a multi-view pose-only estimation approach with its application to GNSS-Visual-Inertial Odometry (GVIO) in this paper. Our main contribution is deriving a visual measurement model which directly associates landmark representation with multiple camera poses and observations. Such a pose-only measurement is proven to be tightly-coupled between landmarks and poses, and maintain a perfect null space that is independent of estimated poses. Finally, we apply the proposed approach to a filter based GVIO with a novel feature management strategy. Both simulation tests and real-world experiments are conducted to demonstrate the superiority of the proposed method in terms of efficiency and accuracy.
\end{abstract}


\end{frontmatter}

\section{Introduction}
High efficiency and accuracy have long been the primary objectives in the fields of Simultaneous Localization and Mapping (SLAM) and vision based multi-sensor fusion. Compared to graph optimization based frameworks, filter based sensor fusion techniques exhibit superior efficiency, enabling pose estimation on platforms with limited computational resources. In recent years, advancements in the Invariant Extended Kalman Filter (IEKF) have significantly improved filtering consistency \citep{barrau2016invariant,barrau2017three,chauchat2024two}, leading to its widespread adoption in various sensor fusion frameworks like GNSS-Visual-Inertial Odometry (GVIO) \citep{liu2023ingvio,xia2024invariant}. 

In vision-aided sensor fusion, the joint estimation of the map and poses significantly increases the computational load, whether using 3D point representations or depth representations. In the context of the IEKF, the tight coupling between map points and poses can lead to poorer computational efficiency, as the state propagation in IEKF is more time consuming \citep{yang2022decoupled}. Consequently, approaches such as Multi-State Constraint Kalman Filter (MSCKF) \citep{mourikis2007multi} and its invariant filtering variants \citep{wu2017vins} have been recognized as a major strategy to mitigate the explosion of state dimensions by marginalizing landmarks from the state before performing measurement update. These methods, however, often feature delayed updates to marginalize landmarks, which result in a less timely state update and potential accumulation of linearization errors \citep{wang2025po,10918798}. Moreover, additional time-consuming operations like anchor frame shift \citep{yang2022decoupled,liu2023ingvio} along with covariance transition are necessary for invariant MSCKF in case the estimated map points may compromise filter consistency. 

In recent years, \cite{wang2025po}, \cite{xueyu2024po} and \cite{xu2025po} have developed the pose-only sensor funsion algorithms based on the pose-only imaging geometry \citep{cai2021pose}, which has shown its potential in enhancing pose accuracy by eliminating the impact of landmark linearization errors. However, these approaches typically rely on two-view geometry whose accuracy depends on the selection of two base frames. And their methods are basically based on traditional EKF or MSCKF, and do not focus on the impact of pose-only model on IEKF.

To address the above issues, we propose a pose-only method that focuses on deriving a closed-form solution for the estimation of pose and visual feature points. This approach allows for the natural elimination of map points during state updates. The proposed method enables instant measurement updates while maintaining filtering consistency. Our contributions can be summarized as follows:
\begin{itemize}
\item We propose a multi-view pose-only representation for visual landmarks, which is shown to be tightly-coupled in terms of the state covariance. 
\item We present a GNSS-Visual-Inertial Odometry with the visual measurement model and a novel feature management strategy based on multi-view pose-only representation, which eliminates the impact from the linearization error of landmarks.
\item We analyze the observability of our proposed measurement model within the invariant filtering framework, demonstrating that it can guarantee the filter consistency without the need to estimate map points.
\end{itemize}

\section{Preliminaries \& Notations}

In the following sections, the coordinate frames and some notations involved are clarified in Table \ref{tab1} and we denote $\boxplus/\boxminus$ as the addition/subtraction operator on manifold. Then the right invariant error for matrix lie group $\mathbf X := (\mathbf R, \mathbf x_1,\cdots, \mathbf x_m), \mathbf R\in SO(3), \mathbf x_1,\cdots,\mathbf x_m \in \mathbb{R}^3$ can be defined:
\begin{equation}
    \begin{aligned}
    \Tilde{\mathbf{X}} = &\mathbf{X}\boxminus \hat{\mathbf{X}}=(\Tilde{\boldsymbol \theta}, \Tilde{\mathbf x}_1 ,\cdots,\Tilde{\mathbf x}_m) \\
    \mathbf{X} = &\Tilde{\mathbf{X}}\boxplus \hat{\mathbf{X}} \\
    = &(exp(\Tilde{\boldsymbol \theta})\hat{\mathbf{R}}, exp(\Tilde{\boldsymbol \theta})\hat{\mathbf{x}}_1+\Tilde{\mathbf x}_1 ,\cdots ,exp(\Tilde{\boldsymbol \theta})\hat{\mathbf{x}}_m+\Tilde{\mathbf x}_m) 
    \end{aligned}\label{eq:3-1}
\end{equation}


\begin{table}[tp]
  \centering
  \caption{Glossary of notation.}
  \setlength{\tabcolsep}{0.8mm}{
  \renewcommand{\arraystretch}{1.9} 
  \begin{tabular*}{250pt}{cccc}
  \hline    
  Symbol & Meaning & Symbol & Meaning \\
  \hline    
  $G$  & Global frame & $I$& IMU Frame \\

  \hline
  $E$ & \makecell[c]{Earth-Centered \\Earth-Fixed Frame} & $C$& Camera Frame\\
  
  \hline
  ${}^B_A\mathbf{R}$ & \makecell[c]{Rotation from frame \\A to frame B} & ${}^B\mathbf{v}_A$&\makecell[c]{Frame A's velocity \\in frame B}\\

  \hline
  \makecell[c]{${}^B\mathbf{p}_A$}
  &\makecell[c]{Frame A's position \\in frame B}
  &$\mathbf{b}_g/\mathbf b_a$&\makecell[c]{Gyroscope/Acceleromter\\ bias}\\ [1pt]
  
  \hline
  $\hat{\mathbf{x}} $ &\makecell[c]{Estimated value of $\mathbf{x}$} & $\Tilde{\mathbf{x}}$  & \makecell[c]{Error state of $\mathbf{x}$} \\ [1pt]
  
  
  
  \hline
  $\lfloor \mathbf{a} \rfloor$  & \makecell{Skew symmetric matrix \\of $\mathbf{a} \in \mathbb{R}^3$} & $exp(\cdot)$ & \makecell[c]{Mapping from $\mathbb{R}^3$ \\to the manifold SO(3)}\\
  
  \hline
  \end{tabular*}}
  \label{tab1}
\end{table}

\section{Multi-view pose-only model}
\subsection{Multi-view feature representation}

Conventional methods adopt a two-view representation in the depth initialization of 3D features \citep{wang2025po,xueyu2024po,xu2025po}. In a sliding window setting, it is more natural to consider a multi-view approach. Assume a 3D feature point $f$, is observed by several camera frames $\{C_i\},i=1,2,...,N$ with the corresponding 2D measurements on the normalized plane $\{\mathbf x_f^{u_i}\}$. We choose $C_1$ as the base frame which is the first camera frame that observes the feature. A two-view imaging geometry equation involving $C_1$ and the $i$-th frame $C_i$ is written as:
\begin{align}
\begin{aligned}
    {}^{C_i}\mathbf p_f &= {}^{C_i}_{C_1}\mathbf R{}^{C_1}\mathbf p_f+{}^{C_i}\mathbf p_{C_1}\\
    d^{C_i}_f\mathbf x^{u_i}_f&= d^{C_1}_f{}^{C_i}_{C_1}\mathbf R\mathbf x^{u_1}_f+{}^{C_i}\mathbf p_{C_1}
\end{aligned}
\end{align}
where ${}^{C_i}\mathbf p_f$ denotes the feature's position in the $i$-th camera's frame. By multiplying $\mathbf x^{u_i}_f$ on both sides of the equation, we can obtain:
\begin{equation}
     d^{C_1}_f\lfloor \mathbf x^{u_i}_f \rfloor {}^{C_i}_{C_1}\mathbf R\mathbf x^{u_1}_f = -\lfloor \mathbf x_f^{u_i} \rfloor {}^{C_i}\mathbf p_{C_1}
\end{equation}
where $d^{C_1}_f$ is the estimated depth in the $C_1$ frame. By stacking all the observations in each frame, the feature depth is initialized by solving a least-squared problem:
\begin{equation}
\begin{aligned}
        \mathbf A_fd_f^{C_1} &= \mathbf b_f, \\
        \mathbf A_f = \begin{bmatrix}
            \lfloor \mathbf x^{u_2}_f \rfloor {}^{C_2}_{C_1}\mathbf R\mathbf x^{u_1}_f \\
            \lfloor \mathbf x^{u_3}_f \rfloor {}^{C_3}_{C_1}\mathbf R\mathbf x^{u_1}_f \\
            \vdots \\
            \lfloor \mathbf x_f^{u_N} \rfloor {}^{C_N}_{C_1}\mathbf R\mathbf x_f^{u_1}
            \end{bmatrix},& \mathbf b_f = \begin{bmatrix}
            -\lfloor \mathbf x_f^{u_2} \rfloor {}^{C_2}\mathbf p_{C_1} \\
            -\lfloor \mathbf x_f^{u_3} \rfloor {}^{C_3}\mathbf p_{C_1} \\
            \vdots \\
            -\lfloor \mathbf x_f^{u_N} \rfloor {}^{C_N}\mathbf p_{C_1}
            \end{bmatrix}
\end{aligned} \label{eq:4-1}
\end{equation}
Therefore, a least-squared solution of the depth is written as:
\begin{equation}
    d^{C_1}_f = (\mathbf A_f^T\mathbf A_f)^{-1}\mathbf A_f^T\mathbf b_f \label{eq:4-2}
\end{equation}
which is the function of $\mathbf X_{obs}=\{{}^G_{C_i}\mathbf R,{}^G\mathbf p_{C_i}\}$ and $\mathbf x_{obs}=\{\mathbf x^{u_i}_f\},i=1,2,...N$. Thus, the global feature position can be written as:
\begin{equation}
\begin{aligned}
    {}^G\mathbf p_f 
    &= d^{C_1}_f{}^G_{C_1}\mathbf R\mathbf x^{u_1}_f+{}^G\mathbf p_{C_1} := g(\mathbf X_{obs}, \mathbf x_{obs})
\end{aligned}\label{eq:4-3}
\end{equation}
This multi-view geometry representation needs only one base frame $C_1$ rather than two candidates used in previous methods \citep{wang2025po,xueyu2024po}. It will be utilized in the following measurement model.

\subsection{Measurement update in a sliding window filter}
Since (\ref{eq:4-3}) has treated the feature as the function of poses and visual observations in the sliding window, the measurement model will not involve the feature Jacobian which is required in conventional EKF and MSCKF:
\begin{equation}
\begin{aligned}
    \mathbf z_f &= \frac{1}{{}^Cz_f}\begin{bmatrix}
        {}^Cx_f \\ {}^Cy_f
    \end{bmatrix} + \mathbf n_f, {}^C\mathbf p_f = \begin{bmatrix}
    {}^Cx_f & {}^Cy_f & {}^Cz_f
\end{bmatrix}^T \\
    \mathbf r_f &= \mathbf z_f - \hat{\mathbf{z}}_f = \mathbf H_{X_{obs}}\Tilde{\mathbf X}_{obs} + \mathbf n_f 
\end{aligned} \label{eq:4-4}
\end{equation}
where $\mathbf n_f\in \mathbb{R}^2$ denotes the visual measurement noise. Here we compare the filter update with the conventional one in MSCKF, and the following proposition can be given:

\begin{prop}
In MSCKF, the poses in the sliding window are updated assuming the feature position is an independent variable with an infinite covariance; while in the pose-only representation, the poses will be updated in a tightly-coupled form. Specifically, the joint covariance of poses and landmarks in traditional MSCKF and our proposed method is given by:
\begin{equation}
\begin{aligned}
    \mathbf P_{MSCKF} &= \begin{bmatrix}
        \mathbf P_X & \mathbf P_{Xf} \\
        \mathbf P_{fX} & \mathbf P_f
    \end{bmatrix}= \begin{bmatrix}
        \mathbf P_X & \mathbf 0 \\
        \mathbf 0 & +\infty
    \end{bmatrix}\\
    \mathbf P_{Ours} &= \begin{bmatrix}
        \mathbf P_X & \mathbf P_{Xf} \\
        \mathbf P_{fX} & \mathbf P_f
    \end{bmatrix}= \begin{bmatrix}
        \mathbf P_{X} & \mathbf P_{X}\mathbf H_g^T \\ \mathbf H_g\mathbf P_{X} & \mathbf H_g\mathbf P_{X}\mathbf H_g^T
    \end{bmatrix}
\end{aligned}
\end{equation}
where $\mathbf H_g = \frac{\partial g}{\partial X}$.

\end{prop}

\section{Application on Invariant GNSS-Visual-Inertial Odometry}
\begin{figure}
    \centering
    \includegraphics[width=0.48\textwidth]{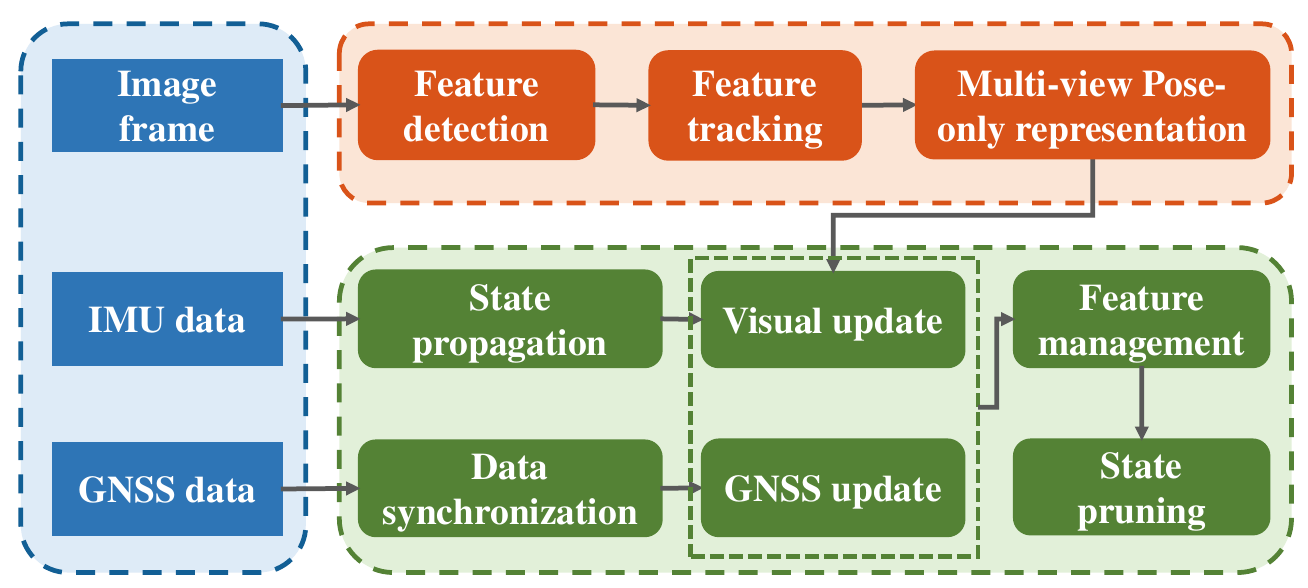}
    \caption{System overview.}
    \label{fig:overview}
\end{figure}

In this section, the multi-view pose only model is applied to an IEKF based tightly-coupled GVIO framework. The system pipeline is illustrated in Fig. \ref{fig:overview}.
\subsection{State definition}
Following the MSCKF framework, we define the state of the system as follows:
\begin{align}
\mathbf{X} &= (
\mathbf{X}_I, \mathbf{X}_{C_1}, ... , \mathbf{X}_{C_N}, d \mathbf{t}_{r}, \dot{d t_{r}}) \label{eq1}\\
\mathbf{X}_I &= (
{ }_{I}^{G}\mathbf{R}, { }^{G}\mathbf{p}_I, {}^G\mathbf{v}_I, \mathbf{b}_g,\mathbf{b}_{a})\label{eq2}\\
\mathbf{X}_C &= ({}^G_C \mathbf{R}, {}^G \mathbf{p}_C)\label{eq:5-1}
\end{align}
where $d\mathbf{t}_{r}$ and $\dot{d t_{r}}$ refer to the GNSS receiver clock bias and corresponding random walk.

\subsection{State propagation}
The filter propagates the state following the classical IMU continuous-time kinematic model and GNSS propagation model given by:
\begin{equation}
    \begin{aligned}
        {}^G_I\dot{\mathbf{R}} &= {}^G_I\mathbf{R}\lfloor {}^I\boldsymbol{\omega}-\mathbf{b}_g-\mathbf{n}_g\rfloor \\
        {}^G\dot{\mathbf{v}}_I &= {}^G_I\mathbf{R}({}^I\mathbf{a}-\mathbf{b}_a-\mathbf{n}_a)+{}^G\mathbf{g} \\
        {}^G\dot{\mathbf{p}}_I &= {}^G\mathbf{v}_I\\
        \dot{\mathbf{b}}_g &= \mathbf{n}_{\omega g}, \dot{\mathbf{b}}_a = \mathbf{n}_{\omega a}\\
        \dot{d \mathbf{t}}_{r,\alpha} &= \dot{d t_{r}}, \alpha = GPS,BDS,GAL,GLO
    \end{aligned} \label{eq:5-2}
\end{equation}
where ${}^G\mathbf g$ is the global gravity vector. By discretizing the continuous-time model, its right invariant error based state transition matrix is written as \citep{yang2022decoupled}:
\begin{equation}
\begin{aligned}
\boldsymbol \Phi_{I_{k,k-1}}&=\begin{bmatrix}
    \mathbf{I}_3 & \mathbf{0}_3 & \mathbf{0}_3 & \boldsymbol \Phi_{14} & \mathbf{0}_3  \\
    \frac{1}{2}\lfloor {}^G \mathbf{g} \rfloor \Delta t_k^2 & \mathbf{I}_3 & \mathbf{I}_3 \Delta t_k & \boldsymbol \Phi_{24} & \boldsymbol \Phi_{25} \\
    \lfloor {}^G \mathbf{g} \rfloor \Delta t_k & \mathbf 0_3 & \mathbf{I}_3 & \boldsymbol \Phi_{34} & \boldsymbol \Phi_{35}\\
    \mathbf{0}_3 & \mathbf{0}_3 & \mathbf{0}_3 & \mathbf{I}_3 & \mathbf{0}_3 \\
    \mathbf{0}_3 & \mathbf{0}_3 & \mathbf{0}_3 & \mathbf{0}_3 & \mathbf{I}_3
    \end{bmatrix} \\
\boldsymbol \Phi_{14} &= -{}^G_{I_{k+1}}\mathbf R\mathbf{J}_r(\Delta \boldsymbol\theta_k), \boldsymbol \Phi_{25} = -{}^G_{I_{k}}\mathbf R\boldsymbol{\Omega}_2 \\
\boldsymbol \Phi_{24} &= -\lfloor {}^G\mathbf p_{I_{k+1}}\rfloor{}^G_{I_{k+1}}\mathbf R\mathbf{J}_r(\Delta \boldsymbol\theta_k)+{}^G_{I_k}\mathbf R\boldsymbol{\Omega}_1 \\
\boldsymbol \Phi_{34} &= -\lfloor {}^G\mathbf v_{I_{k+1}}\rfloor{}^G_{I_{k+1}}\mathbf R\mathbf{J}_r(\Delta \boldsymbol\theta_k)+{}^G_{I_k}\mathbf R\boldsymbol{\Omega}_3 \\
\boldsymbol \Phi_{35} &= -{}^G_{I_{k}}\mathbf R\boldsymbol{\Omega}_4
\end{aligned} \label{eq:5-3}
\end{equation}
where $\Delta \boldsymbol \theta_k = exp({}^I\boldsymbol\omega\Delta t_k)$, $\mathbf J_r(\cdot)$ is the right Jacobian for $SO(3)$ group, $\boldsymbol{\Omega}_{i}, i=1,2,3,4$ are time-varying blocks related to IMU biases. 
 
\subsection{Pose-only visual update}
If a feature $f$ is observed by more than three camera frames $\mathcal{C}_f=\{C_l\}, l=i,...,j$ where $C_i$ is the first camera pose observing $f$, we initialize the feature depth through (\ref{eq:4-1}). In the update step, when $f$ is projected onto the $k$-th camera pose, the measurement model is defined as follows:
\begin{equation}
\begin{aligned}
    \mathbf{x}^{u_k}_f &= h(\mathbf{X}_{C_k},{}^G\mathbf{p}_f) +\mathbf n^{u_k} = \frac{1}{{}^{C_k}\mathbf{z}_f}\begin{bmatrix}
    {}^{C_k}\mathbf{x}_f \\ {}^{C_k}\mathbf{y}_f
    \end{bmatrix}+\mathbf n^{u_k} \\
    {}^G\mathbf p_f &= g(\mathbf X, \mathbf x_f), \mathbf x_f = (\mathbf x^{u_i}_f, \cdots, x^{u_{j}}_f)
\end{aligned} \label{eq:5-7}
\end{equation}
where $\mathbf x_f$ denotes the set of observations. By linearizing the projection equation, the corresponding measurement Jacobians are written as:
\begin{equation}
\begin{aligned}
    \mathbf H_{\mathbf X_C} &= \begin{bmatrix}
        \cdots \mathbf H_{\mathbf X_{C_m}}\cdots    \end{bmatrix}, \mathbf H_n = \begin{bmatrix}
        \cdots \mathbf H_{n_m} \cdots     \end{bmatrix} \\
    \mathbf H_{\mathbf X_{C_m}} &= \left\{\begin{array}{ll}
         \frac{\partial h}{\partial {}^G \mathbf p_f}\frac{\partial g}{\partial \mathbf X_{C_m}}, m \neq k \\
        \frac{\partial h}{\partial \mathbf X_{C_m}} + \frac{\partial h}{\partial {}^G \mathbf p_f}\frac{\partial g}{\partial \mathbf X_{C_m}}, m=k
    \end{array}\right. \\
    \mathbf H_{n_m} &= \left\{\begin{array}{ll}
         \frac{\partial h}{\partial {}^G \mathbf p_f}\frac{\partial g}{\partial \mathbf n^{u_m}},m \neq k \\
         \mathbf I_2 + \frac{\partial h}{\partial {}^G \mathbf p_f}\frac{\partial g}{\partial \mathbf n^{u_m}}, m = k
    \end{array}\right. \\
    \frac{\partial h}{\partial {}^G \mathbf p_f} &= \mathbf J_\pi{}^{G}_{C_k}\mathbf R^T,
    \frac{\partial h}{\partial \mathbf X_{C_k}} = -\mathbf J_\pi\begin{bmatrix}
        \mathbf 0_3 & {}^{G}_{C_k}\mathbf R^T
    \end{bmatrix} \\
    \mathbf J_\pi &= \frac{1}{{}^{C_k}\mathbf{z}^2_f}\begin{bmatrix}
        {}^{C_k}\mathbf{z}_f & 0 & -{}^{C_k}\mathbf{x}_f \\
        0 & {}^{C_k}\mathbf{z}_f & -{}^{C_k}\mathbf{y}_f
    \end{bmatrix}
\end{aligned} \label{eq:5-8}
\end{equation}
Then the regular EKF update is performed without the null space projection in MSCKF.

\subsection{Feature management}
\begin{figure}[t]
    \centering
    \includegraphics[width=0.45\textwidth]{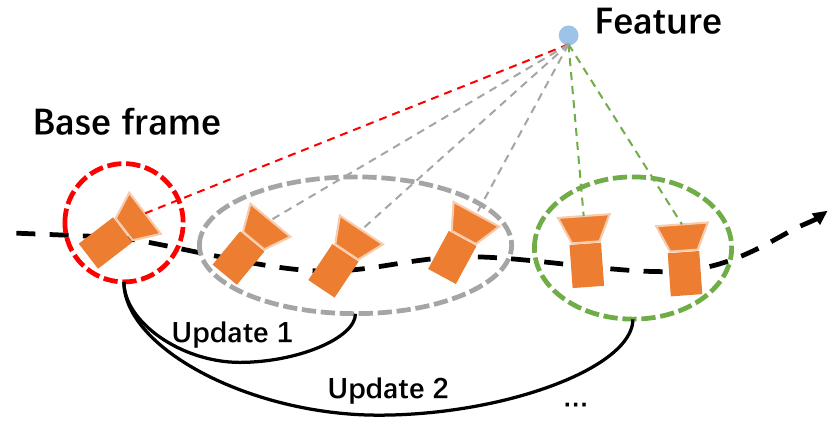}
    \caption{Illustration of pose only feature update. The observation of the first frame (base frame) is involved in the first round update with other camera frames (gray). The observation on the base frame continues to be used in the subsequent updates with new multiple camera frames (green). }
    \label{fig:feature_update}
\end{figure}
The detected feature will be initialized when it has collected more than three observations. Once the visual update is completed, the initialized feature and the associated measurements except the measurement in the base frame will be deleted from the map. If new observations are available, the base frame will remain in the sliding window and reused for the new initialization process, as shown in Fig. \ref{fig:feature_update}; if the feature is no longer tracked or the base frame is ready to be marginalized, the base frame observation will be deleted. And since pose-only features do not depend on the anchored camera frame, operations such anchor change in invariant MSCKF \citep{liu2023ingvio} are not required when the base frame is marginalized.

\subsection{GNSS update} \label{sec:left_update}
The GNSS pseudorange $P$ and Doppler observation $D$ between receiver $r$ and satellite $s$ can be expressed simply by the following:
\begin{equation}
\begin{aligned}
P &= \rho_{r}^{s}+c d t_{r}-c d t^{s}+Tro+Ion+\varepsilon_{P} \\
D &= \frac{1}{\lambda}(\dot{\rho}_{r}^{s}+
      c\dot{dtr}-c\dot{dt^{s}})+
      \varepsilon_D  
\end{aligned} \label{eq:5-4}
\end{equation}
where $\rho_{r}^{s}$ represents the true geometric distance. The symbol $c$ denotes the speed of light. The term $dt^{s}$ indicates the clock bias of satellite $s$, which can be eliminated using precise clock products. $Tro$ and $Ion$ signify the tropospheric and ionospheric delay, respectively. $\lambda$ is the wavelength, and $\varepsilon$ stands for noise. GNSS updates are performed once the GNSS measurements become available.

\section{Pose-only Observability Analysis}
In this part, we investigate the observability property of the proposed measurement model. The analysis is based on the observability matrix used in previous literature \citep{li2013high}. The estimated state at time $t_k$ is defined as follows:
\begin{equation}
    \mathbf X_k = (\mathbf X_{I_k}, \mathbf X_{C_i}, \mathbf X_{C_{i+1}}, \cdots, \mathbf X_{C_{k-1}}) \label{eq:4-7}
\end{equation}
where $\mathbf X_{C_{l}}, l=i,...,k-1$ and $\mathbf X_k$'s cloned state $\mathbf X_{C_{k}}$ are involved in the initialization of the feature. The oldest frame $C_i$ is assumed to be the base frame. Following the right invariant error definition, the state transition matrix from $t_0$ to $t_k$ is derived as:
\begin{equation}
\begin{aligned}
\boldsymbol{\Phi}_{k,0}
&=\left[\begin{array}{cccccc}
\boldsymbol{\Phi}_{I_{k,0}} & \mathbf{0}_{15\times 6} & \cdots & \mathbf{0}_{15\times 6} \\
\boldsymbol{\Phi}_{C_iI,0} & \boldsymbol{\Phi}_{C_{i,0}} & \cdots & \mathbf{0}_{15\times 6} \\
\vdots & \vdots & \ddots & \vdots \\
\boldsymbol{\Phi}_{C_{k-1}I,0} & \mathbf{0}_{6\times 15} & \cdots & \boldsymbol{\Phi}_{C_{k-1,0}} 
\end{array}\right] \\
\boldsymbol{\Phi}_{C_l,0} &= \begin{bmatrix}
    \mathbf{I}_3 & \mathbf{0}_3 \\
    \frac{1}{2}\lfloor {}^G \mathbf{g} \rfloor \Delta t_{l}^2 & \mathbf{I}_3
\end{bmatrix}
\end{aligned}  \label{eq:4-8}
\end{equation} 
The $k$-th row block of the final observability matrix $\mathcal{O}$ can be derived as:
\begin{equation}
\begin{aligned}
    \mathcal{O} &\triangleq\begin{bmatrix}
    \mathbf{H}_{0}^T & (\mathbf{H}_{1}\boldsymbol{\Phi}_{1,0})^T & \cdots & (\mathbf{H}_{m}\boldsymbol{\Phi}_{m,0})^T
    \end{bmatrix}^T \\
    \mathcal{O}_{k} &= \mathbf{H}_k\boldsymbol{\Phi}_{k,0} \\
    &= \begin{bmatrix}
        \mathbf H_{I_k,\theta} \,\, \mathbf H_{I_k,p} \,\, \mathbf 0_{2\times 9} \,\, \mathbf H_{C_i,\theta} \,\, \mathbf H_{C_i,p}&\cdots
    \end{bmatrix}\boldsymbol\Phi_{k,0} \\
    &= \begin{bmatrix}
        \mathbf \Gamma_{I_k} & \mathbf \Gamma_{C_i} & \cdots & \mathbf \Gamma_{C_{k-1}}
    \end{bmatrix} \\
    \mathbf \Gamma_{I_k} &= \begin{bmatrix}
        \mathbf H_{I_k,\theta}+\frac{1}{2}\mathbf H_{I_k,p}\lfloor {}^G \mathbf{g} \rfloor \Delta t_k^2  & \, \mathbf H_{I_k,p} \, & \mathbf \Gamma_{I_k,*}
    \end{bmatrix} \\
    \mathbf \Gamma_{C_l} &= \begin{bmatrix}
        \mathbf H_{C_l,\theta}+\frac{1}{2}\mathbf H_{C_l,p}\lfloor {}^G \mathbf{g} \rfloor \Delta t_l^2  & \, \mathbf H_{C_l,p}
    \end{bmatrix}
\end{aligned}
\end{equation}
where $\mathbf H_k$ is the measurement Jacobian for the measurement model in (\ref{eq:5-7}), $\mathbf \Gamma_{I_k,*}$ represents the matrix block which will not affect the analysis. It is easy to verify that $\mathbf H_k$ satisfies the rotational and translational constraint:
\begin{equation}
\begin{aligned}
    \mathbf H_k &= \begin{bmatrix}
        \mathbf H_{C_k} & \mathbf H_{C_i} & \cdots & \mathbf H_{C_{k-1}}
    \end{bmatrix} \\
    \sum\limits_{l=i}^k \mathbf H_{C_l,\theta}&=\mathbf 0_{2\times 3},\sum\limits_{l=i}^k \mathbf H_{C_l,p}=\mathbf 0_{2\times 3}
\end{aligned}
\end{equation}
where $\mathbf H_{Ci}=[\mathbf H_{C_i,\theta} \quad \mathbf H_{C_i,p}]$. Therefore, the null space can be derived as:
\begin{equation}
    \mathcal{N}_{po} = 
    \setlength{\arraycolsep}{1pt}
    \begin{bmatrix}
    {}^G\mathbf{g}^T & \mathbf{0}_{1\times3} & \mathbf{0}_{1\times9} & {}^G\mathbf{g}^T & \mathbf{0}_{1\times3} & \cdots & {}^G\mathbf{g}^T & \mathbf{0}_{1\times3} \\
    \mathbf{0}_3 & \mathbf{I}_3 & \mathbf{0}_{3\times 9} &  \mathbf{0}_3 &\mathbf{I}_3 & \cdots & \mathbf{0}_3 & \mathbf{I}_3 
    \end{bmatrix}^T \label{eq:4-9}
\end{equation}
It is clear from (\ref{eq:4-9}) that the null space for the pose-only filter is always independent of the estimated state.

\section{EXPERIMENT AND ANALYSIS} \label{sec:evaluation}
We conduct both simulations and real-world tests to validate the consistency and pose accuracy of our proposed method. In real-world tests, we compare the positional accuracy of our method with that of state-of-the-art GVIO algorithms.
\begin{figure}[t]
\centering
\includegraphics[scale=0.6]{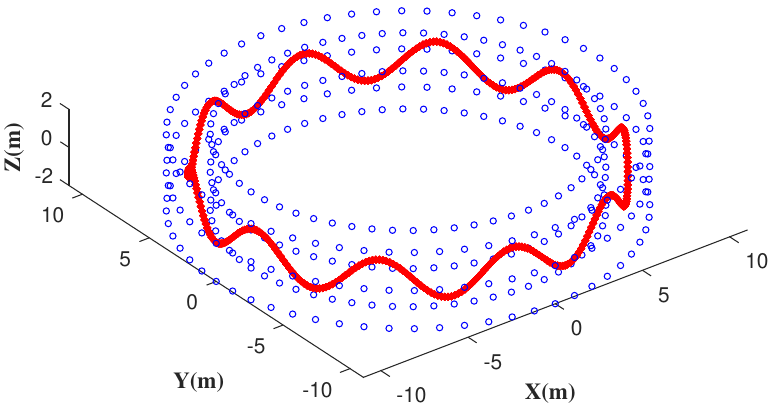}
\caption{The simulated trajectory (red) and landmarks (blue).}
\label{fig:sim_scenario}
\end{figure}
\subsection{Simulation}

\begin{table}[h]
    \caption{RMSE (deg/m) and NEES of orientation and position for 30 Monte Carlo runs.}
    \label{tab:sim_nees_rmse}
    \centering
    \renewcommand\arraystretch{1.2}
    \begin{tabular}{ccccc}
        \hline
        \makecell{Pixel Noise} & 
        \makecell[l]{Algorithm} & \makecell{RMSE \\ Ori.} & \makecell{RMSE \\ Pos.} & \makecell{NEES \\ Pose} \\
        \hline
        \multirow{4}*{1} & 
        \makecell[l]{InVIO(L=5)} & \makecell{1.537} & \makecell[c]{0.289} & \makecell{6.705} \\
        & \makecell[l]{InVIO(L=10)} & \makecell{1.072} &
        \makecell[c]{0.202} & \makecell{6.167} \\
        & \makecell[l]{Po-InVIO(two)} & \makecell{1.157} & \makecell[c]{0.212} & \makecell{6.313} \\
        & \makecell[l]{Po-InVIO(multi)} & \makecell{0.832} & \makecell[c]{0.158} & \makecell{5.958} \\
        \hline
        \multirow{4}*{2} & 
        \makecell[l]{InVIO(L=5)} & \makecell{1.419} & \makecell[c]{0.305} & \makecell{6.071} \\
        & \makecell[l]{InVIO(L=10)} & \makecell{1.278} &
        \makecell[c]{0.255} & \makecell{6.575} \\
        & \makecell[l]{Po-InVIO(two)} & \makecell{1.215} & \makecell[c]{0.237} & \makecell{6.235} \\
        & \makecell[l]{Po-InVIO(multi)} & \makecell{1.197} & \makecell[c]{0.230} & \makecell{6.276} \\
        \hline
    \end{tabular}
\end{table}

\begin{figure}
    \centering
    \includegraphics[scale=0.55]{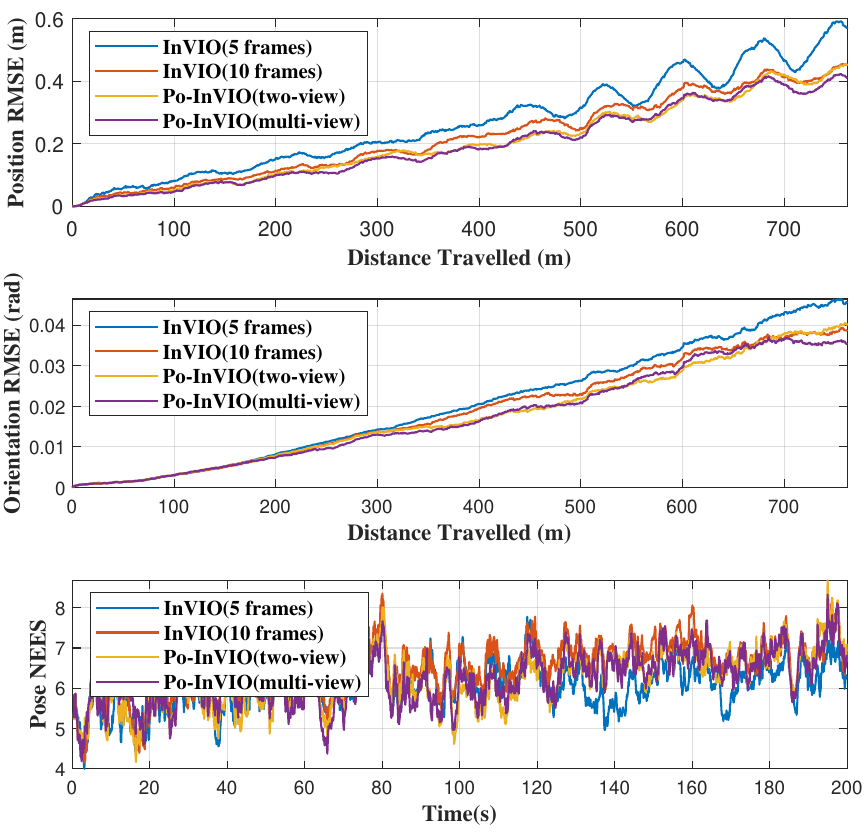}
\caption{Pose RMSE and NEES for 30 Monte Carlo runs where pixel noise is 2 px.}
\label{fig:sim_nees_rmse}
\end{figure}

In the simulation test, we create the ground truth trajectory lasting 200 seconds and sensor measurements utilizing a 100 Hz IMU and a 10 Hz camera, as shown in Fig. \ref{fig:sim_scenario}. The continuous-time IMU gyroscope and accelerometer white noise is $3\times 10^{-4}$ (unit: $rad\cdot s^{-0.5}$) and $1\times 10^{-3}$ (unit: $m\cdot s^{-1.5}$) respectively, and the random walk is set to $3\times 10^{-5}$ (unit: $rad\cdot s^{-1.5}$) and $2\times 10^{-4}$ (unit: $m\cdot s^{-2.5}$). To illustrate the performance of pose-only representation, we compare the proposed pose-only VIO (Po-InVIO) with the conventional right invariant MSCKF based VIO (InVIO). Po-InVIO performs the update once the feature is tracked for no less than 4 frames. As traditional MSCKF based InVIO requires a threshold $L$ for the number of observing camera frames in feature initialization, we test two cases including $L=5$ and $L=10$. We also compare the effect between the two-view representation and multi-view representation. To be specific, the two-view representation selects the first and last observation for each feature as two base frames, and the feature depth is represented in the first camera frame that observes the feature. The pose accuracy and consistency are evaluated by Root Mean Squared Error (RMSE) and Normalized Estimation Error Squared (NEES) for 30 Monte Carlo runs. 


The results are shown in Fig. \ref{fig:sim_nees_rmse} and Table \ref{tab:sim_nees_rmse}. The accuracy of traditional MSCKF is observed to depend on the number of tracking frames for accurate feature initialization. The pose-only representation improves pose estimation by adopting a tightly coupled measurement update model along with the proposed feature management strategy shown in Fig. \ref{fig:feature_update}. Furthermore, by utilizing multiple frames to represent features, our approach is slightly better than two-view based representation in terms of pose accuracy. Since our proposed system guarantees a perfect null space that is independent of the feature position, the NEES is close to the ideal value 6, demonstrating a good consistent estimation. 

\subsection{Real-world Experiment}
\begin{table}[t]
    \caption{Absolute Translation Error (m) of compared VIO algorithms on EuRoC Dataset.}
    \label{tab2}
    \centering
    \renewcommand\arraystretch{1.2}
    \begin{tabular}{cccccc}
        \hline
        \multirow{2}*{Seq} & \multicolumn{5}{c}{Algorithm} \\
        \Xcline{2-6}{0.2pt}& \makecell[c]{VINS-\\Mono}  & \makecell[c]{OpenVINS} & \makecell[c]{POCKF} & \makecell[c]{Ours\\(w/o po)} & \makecell[c]{Ours} \\
        \hline
        
        \makecell[c]{MH\_01} & \makecell[c]{0.167} & \makecell[c]{0.144} & \makecell[c]{0.135} & \makecell[c]{0.158}  & \makecell[c]{\textbf{0.110}} \\

        \makecell[c]{MH\_02} & \makecell[c]{0.181} & \makecell[c]{0.154} & \makecell[c]{0.164} & \makecell[c]{0.146}  & \makecell[c]{\textbf{0.133}} \\
        
        \makecell[c]{MH\_03} & \makecell[c]{\textbf{0.185}} & \makecell[c]{0.269} & \makecell[c]{0.231} & \makecell[c]{0.226}  & \makecell[c]{0.220}  \\
        
        \makecell[c]{MH\_04} & \makecell[c]{0.369} & \makecell[c]{\textbf{0.196}} & \makecell[c]{0.202} & \makecell[c]{0.200} & \makecell[c]{0.214}  \\
        
        \makecell[c]{MH\_05} & \makecell[c]{0.302} & \makecell[c]{0.333} & \makecell[c]{0.278} & \makecell[c]{0.329} & \makecell[c]{\textbf{0.273}} \\
        
        \makecell[c]{V1\_01} & \makecell[c]{0.083} & \makecell[c]{0.065} & \makecell[c]{0.071} & \makecell[c]{0.057}  & \makecell[c]{\textbf{0.052}}  \\

        \makecell[c]{V1\_02} & \makecell[c]{0.118} & \makecell[c]{0.077} & \makecell[c]{0.179} & \makecell[c]{0.077}  & \makecell[c]{\textbf{0.053}} \\

        \makecell[c]{V1\_03} & \makecell[c]{0.187} & \makecell[c]{\textbf{0.071}} & \makecell[c]{0.207} & \makecell[c]{0.074}  & \makecell[c]{0.083} \\

        \makecell[c]{V2\_01} & \makecell[c]{0.086} & \makecell[c]{0.121} & \makecell[c]{\textbf{0.077}} & \makecell[c]{0.115}  & \makecell[c]{0.100} \\

        \makecell[c]{V2\_02} & \makecell[c]{0.149} & \makecell[c]{\textbf{0.079}} & \makecell[c]{0.203} & \makecell[c]{0.106}  & \makecell[c]{0.090}  \\

        \makecell[c]{V2\_03} & \makecell[c]{0.335} & \makecell[c]{0.204} & \makecell[c]{0.265} & \makecell[c]{\textbf{0.193}}  & \makecell[c]{0.218} \\

        \hline

        \makecell[c]{Average} & \makecell[c]{0.217} & \makecell[c]{0.176} & \makecell[c]{0.193} & \makecell[c]{0.171}  & \makecell[c]{\textbf{0.158}} \\
        \hline
    \end{tabular}
\end{table}

\begin{figure}[ht]
\centering
\includegraphics[scale=0.6]{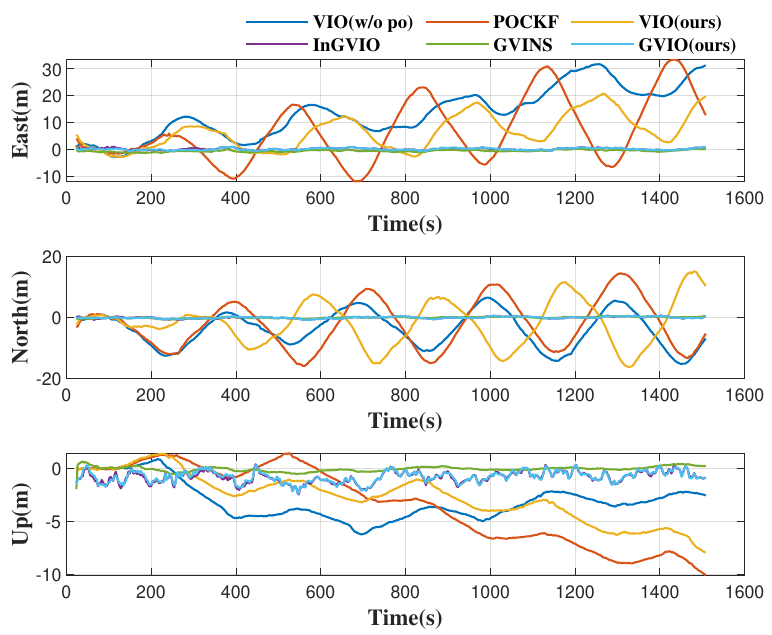}
\caption{ENU error of compared algorithms on the \textit{sports}\_\textit{field} sequence.}
\label{fig:error_sf}
\end{figure}

\begin{figure}[ht]
\centering
\includegraphics[scale=0.6]{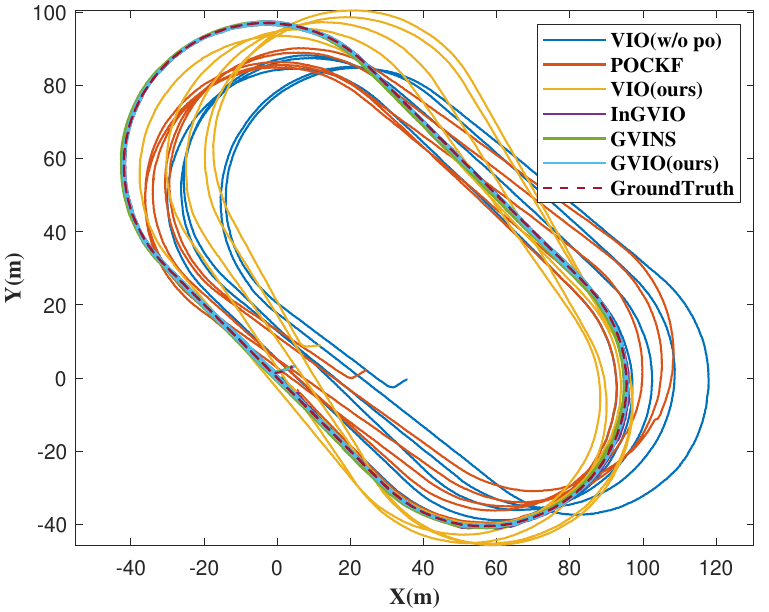}
\caption{The trajectories of compared algorithms on the \textit{sports}\_\textit{field} sequence.}
\label{fig:traj_sf}
\end{figure}
\subsubsection{VIO test}
We validate our algorithm on VIO algorithms using the public EuRoC Dataset \citep{burri2016euroc}. The dataset consists of both visual data captured from a stereo 20Hz camera and 200Hz IMU. Data collection was carried out in various settings, including a factory and a small indoor environment under different conditions, such as dynamic lighting and varying levels of motion blur. The open source algorithms OpenVINS \citep{geneva2020openvins} and VINS-Mono \citep{qin2018vins} are compared as two benchmarks. To illustrate the impact of the proposed pose-only model, we also test the performance of traditional right invariant MSCKF (ours(w/o po)). And to compare the difference between two-view and multi-view representation, we implement a pose-only constraint VIO termed POCKF based on the two-view representation \citep{cai2021pose} with delayed update as traditional MSCKF. The sliding window size for filter based VIO is set to 25, and 150 features are detected in the visual frontend.

Table \ref{tab2} reports the Absolute Translation Error (ATE) on the tested sequences. Thanks to the multi-view representation and the update strategy, our proposed VIO estimates the pose in a more immediate and robust way. It results in a better performance especially in some challenging sequences like MH\_05 sequence with rapid motion and low light conditions. In some sequences with rapid rotation motion patterns such as V1\_03 and V2\_03, the depth representation using fewer image observations than that in conventional MSCKF leads to the unstable estimation accuracy for POCKF. However, our method achieves a robust performance on average because of the multi-view representation that models the feature depth more accurately. 
\subsubsection{GVIO test}
\begin{table}[t]
    \caption{Absolute Translation Error (m)/Average processing time per frame (ms) on the GVINS Dataset.}
    \label{tab:GVINS_dataset_APE}
    \centering
    \renewcommand\arraystretch{1.5}
    \setlength\tabcolsep{2.5pt}
    \begin{tabular}{cccc}
        \hline
        Algorithm  & \textit{sports\_field} & \textit{complex\_environment} & \textit{urban\_driving} \\
        \hline

        \makecell[c]{GVINS} & \makecell[c]{\textbf{0.76}/30.42} & \makecell[c]{3.57/23.49} & \makecell[c]{5.67/40.84}  \\
        
        \makecell[c]{InGVIO} & \makecell[c]{1.07/7.39} & \makecell[c]{2.69/6.34} & \makecell[c]{5.54/5.36}  \\
        


        \makecell[c]{POCKF} & \makecell[c]{15.96/5.45} & \makecell[c]{29.32/4.30} & \makecell[c]{$\times$}   \\

        \makecell[c]{VIO(w/o po)} & \makecell[c]{18.41/5.03} & \makecell[c]{33.02/4.52} & \makecell[c]{$\times$}   \\
        
        \makecell[c]{VIO(ours)} & \makecell[c]{13.30/5.79} & \makecell[c]{25.86/4.23} & \makecell[c]{$\times$}   \\

        \makecell[c]{GVIO(ours)} & \makecell[c]{1.06/7.52} & \makecell[c]{\textbf{2.60}/6.27} & \makecell[c]{\textbf{5.48}/5.48}   \\

        \hline
    \end{tabular}
\end{table}
In the GVIO experiment, we test the algorithm based on the GVINS dataset \citep{cao2022gvins}. The GVINS dataset is collected by a helmet with a visual inertial sensor suite and an u-blox ZED-F9P GNSS receiver. The rtk engine embodied in the GNSS receiver provides the positioning ground truth. The dataset includes three long-distance sequences: \textit{sports}\_\textit{field}, \textit{complex\_environment} with indoors and outdoor scenarios, and \textit{urban\_driving}. Each sequence has lasted for more than twenty minutes with GNSS limited regions. We compare our algorithms with the open-source algorithms including GVINS \citep{cao2022gvins} and InGVIO \citep{liu2023ingvio}. The VIO algorithms adopting the pose-only model (VIO(ours)) and VIO with the traditional MSCKF measurement model (VIO(w/o po)), are also tested to demonstrate the performance of multi-view pose-only model. Since VIO cannot be run through the \textit{urban\_driving} sequence, we only compare its performance in the other two sequences. The first 5\% of VIO output poses is aligned with the ground truth to obtain a global trajectory in ECEF.

The statistics for positioning accuracy are shown in Table \ref{tab:GVINS_dataset_APE}. In the outdoor scenarios, pose-only VIO can achieve a substantial improvement in accuracy compared to the conventional VIO, which is illustrated in Fig. \ref{fig:error_sf} and Fig. \ref{fig:traj_sf}. Specifically, POCKF performs slightly better than traditional MSCKF and the proposed multi-view based pose-only VIO further performs better than POCKF. When our method is used in GVIO, the estimation accuracy of different algorithms is similar because of the relatively accurate GNSS observations. 

\bibliography{ifacconf}             
\end{document}